\documentclass[12pt]{article}
\usepackage{amsmath}
\usepackage{times}
\usepackage{graphicx}
\usepackage{color}
\usepackage{multirow}
\usepackage{url} 
\usepackage{caption}
\usepackage{geometry}
\usepackage{floatrow} 
\usepackage{algorithm}
\usepackage{algorithmic}
\usepackage[authoryear]{natbib}
\usepackage{rotating}
\usepackage{bbm}
\usepackage{latexsym}
\usepackage{amssymb}
\usepackage{subcaption}

\newcommand{\captionfonts}{\normalsize}

\makeatletter  
\long\def\@makecaption#1#2{%
  \vskip\abovecaptionskip
  \sbox\@tempboxa{{\captionfonts #1: #2}}%
  \ifdim \wd\@tempboxa >\hsize
    {\captionfonts #1: #2\par}
  \else
    \hbox to\hsize{\hfil\box\@tempboxa\hfil}%
  \fi
  \vskip\belowcaptionskip}
\makeatother   

\begin{document}
\hspace{13.9cm}1

\ \vspace{20mm}\\

{\LARGE Boosting MCTS with Free Energy Minimization}

\ \\
{\bf \large Mawaba Pascal Dao$^{\displaystyle 1}$, Adrian M. Peter$^{\displaystyle 1}$} \\
{$^{\displaystyle 1}$Florida Institute of Technology}\\
%


\thispagestyle{empty}
\markboth{}{NC instructions}
\ \vspace{-0mm}\\
%
\begin{center}
{\bf Abstract}
\end{center}
Active Inference, grounded in the Free Energy Principle, provides a powerful lens for understanding how agents balance exploration and goal-directed behavior in uncertain environments. Here, we propose a new planning framework, that integrates Monte Carlo Tree Search (MCTS) with active inference objectives to systematically reduce epistemic uncertainty while pursuing extrinsic rewards. Our key insight is that MCTS\textemdash already renowned for its search efficiency\textemdash can be naturally extended to incorporate free energy minimization by blending expected rewards with information gain. Concretely, the Cross-Entropy Method (CEM) is used to optimize action proposals at the root node, while tree expansions leverage reward modeling alongside intrinsic exploration bonuses. This synergy allows our planner to maintain coherent estimates of value and uncertainty throughout planning, without sacrificing computational tractability. Empirically, we benchmark our planner on a diverse set of continuous control tasks, where it demonstrates performance gains over both stand-alone CEM and MCTS with random rollouts.


\section{Introduction}

The integration of search mechanisms into decision-making frameworks has consistently led to significant performance improvements across various domains. Monte Carlo Tree Search (MCTS), a powerful search-based planning method, has been particularly successful in discrete domains such as game-playing, with notable applications like AlphaGo combining MCTS with deep neural networks to achieve superhuman performance in the game of Go \citep{silver2016mastering, silver2017mastering}. However, extending MCTS to more general settings, particularly within the Active Inference framework, presents both challenges and opportunities.

Active Inference, rooted in the Free Energy Principle \citep{friston2010free}, provides a unifying framework for understanding action and perception as processes of minimizing free energy. Recent advancements have explored the integration of Active Inference with MCTS to enable sophisticated planning under uncertainty in both discrete and continuous state-action spaces \citep{fountas2020deep, tschantz2020reinforcement}. These methods demonstrate the potential to balance exploitation and exploration naturally by incorporating free energy minimization as a criterion for action selection. However, key challenges remain, including the computational demands of planning in continuous spaces, ensuring reliable value estimation during tree-based search, and extending these methods to practical applications and benchmarks beyond example problems.

In this paper, we propose a novel framework that integrates MCTS with Active Inference to address these challenges. Our approach introduces mechanisms for efficient planning in continuous state-action spaces while aligning the generative model of Active Inference with the tree search process.

Our contributions can be summarized as follows:

\begin{itemize}    
    \item \textbf{Root Action Distribution Planning:} We propose a novel mechanism where a single Gaussian action distribution is fitted at the root node using the Cross-Entropy Method (CEM). This root action distribution is utilized consistently throughout the tree traversal and simulation phases, significantly reducing computational complexity while ensuring value estimation remains aligned with actual action selection. By constraining the tree size, we maintain the validity of the root action distribution, enabling efficient and reliable planning.
    
    \item \textbf{Enhanced Exploration through Information Gain:} Our method incorporates intrinsic exploration by integrating epistemic value (Information Gain) into the planning process. This dual exploration mechanism, achieved through both the expected free energy criterion and MCTS exploration, improves the agent's ability to navigate high-dimensional continuous domains.
\end{itemize}

The remainder of this paper is organized as follows. In Section 2, we discuss related work, providing a comprehensive overview of existing methods that integrate MCTS with Active Inference. Section 3 provides a background on MCTS, the Free Energy Principle, and Active Inference. Section 4 describes our proposed planner in detail, explaining each component and its mathematical grounding. Section 5 presents experimental results demonstrating the effectiveness of our approach.  Finally, in the conclusion, we discuss the broader implications of our findings and highlight promising directions for future research.

\section{Related Work}

Active Inference, rooted in the Free Energy Principle \citep{friston2010free}, offers a unified framework for understanding perception and action as inference processes. It has demonstrated broad applicability across neuroscience and machine learning, modeling phenomena such as curiosity \citep{schwartenbeck2018computational}, dopaminergic discharges \citep{fitzgerald2015dopamine}, and animal navigation. However, a significant challenge lies in the computational complexity of evaluating all possible policies, which grows exponentially with the planning horizon.

Model-based Reinforcement Learning (RL) methods aim to learn a model of the environment's dynamics and use it for planning \citep{moerland2023model}. These methods can be more sample-efficient than model-free approaches, as they can simulate experiences without interacting with the environment. \citet{chua2018deep} proposed PETS (Probabilistic Ensembles with Trajectory Sampling), which uses an ensemble of probabilistic models for planning in continuous action spaces. Similarly, \citet{hafner2019learning} introduced PlaNet, a model-based RL method that learns a latent dynamics model for planning. By leveraging probabilistic models, these methods provide robust uncertainty quantification, which is critical for exploration and planning under uncertainty.

\citet{tschantz2020reinforcement} proposed the Free Energy of Expected Future (FEEF) as a tractable objective for decision-making in RL environments. Their method incorporates a model-based Cross-Entropy Method (CEM) for policy optimization, achieving a balance between exploration and exploitation in sparse and continuous control tasks.

Monte Carlo Tree Search (MCTS), a decision-making framework, has proven valuable in addressing the computational complexity of evaluating an exponentially growing number of possible policies as the planning horizon increases. MCTS achieves this by sampling a subset of possible policies. Early applications of MCTS focused on discrete domains, such as game playing \citep{coulom2006efficient}, with significant successes in AlphaGo \citep{silver2016mastering, silver2017mastering}. While extensions to continuous action spaces, such as progressive widening \citep{coulom2006efficient} and hierarchical optimization \citep{bubeck2011x}, have broadened its scope, these approaches are typically employed in Reinforcement Learning (RL) contexts.

Recent advancements have combined MCTS with Active Inference to address challenges in planning under uncertainty. For instance, \citet{fountas2020deep} proposed an MCTS-based Active Inference framework that replaces traditional selection criteria, such as the Upper Confidence Bounds applied to Trees (UCT) \citep{kocsis2006bandit}, with an expected free energy (EFE)-based criterion. Their approach employs a deep neural network to approximate posterior distributions and utilizes Monte Carlo sampling to evaluate free energy terms efficiently. This integration demonstrated improved performance in tasks requiring sophisticated planning, such as the dSprites dataset and the Animal-AI environment. 

Branching Time Active Inference (BTAI), introduced by \citet{champion2022branching}, further unified MCTS and Active Inference by framing planning as Bayesian model expansion. BTAI treats the tree structure as part of the generative model itself, dynamically expanding the model to incorporate future observations and latent variables. This approach reduced the computational overhead associated with traditional Active Inference models, allowing applications in graph navigation and other complex tasks.

Despite these advances, the integration of MCTS with the Free Energy Principle in practical applications remains underexplored. Most implementations adopt MCTS as a planning tool within RL frameworks, leaving the potential of Active Inference—particularly its capacity for intrinsic motivation and uncertainty minimization—relatively untapped.

In this work, we address these gaps by integrating MCTS with Active Inference in a novel way. Our framework employs a model-based approach, using MCTS for planning EFE-optimal paths in continuous state-action spaces. We build upon methods such as progressive widening and ensemble modeling to extend MCTS to continuous domains while maintaining compatibility with the generative model of Active Inference. This approach enables efficient exploration and exploitation in environments characterized by high uncertainty and sparse rewards.

\section{Background}

\subsection{Active Inference and the Free Energy Principle}

The Free Energy Principle, originating in neuroscience, posits that systems act to minimize a quantity called free energy, which measures how well an internal generative model predicts observations \citep{friston2010free}. This principle unifies perception and action under the framework of probabilistic inference, where agents aim to align their beliefs with observed data and predict future states.

Free energy is defined as:

\begin{equation}
F(Q, y) = D_{\text{KL}}[Q(x) \| P(x|y)] - \ln P(y),
\end{equation}

where:
\begin{itemize}
  \item \( Q(x) \): The approximate posterior distribution over hidden states \( x \).
  \item \( P(x|y) \): The true posterior distribution over hidden states, given observations \( y \).
  \item \( \ln P(y) \): The log evidence (marginal likelihood) of the observations.
  \item \( D_{\text{KL}}[Q(x) \| P(x|y)] \): The Kullback-Leibler divergence between the approximate and true posterior distributions.
\end{itemize}

Minimizing free energy involves two components:
\begin{enumerate}
    \item Reducing the Kullback-Leibler (KL) divergence, which aligns the approximate posterior \( Q(x) \) with the true posterior \( P(x|y) \).
    \item Maximizing the log evidence \( \ln P(y) \), which ensures that the generative model predicts observations accurately.
\end{enumerate}

Building on this principle, Active Inference provides a framework for decision-making, where agents minimize variational free energy by simultaneously improving their beliefs about the environment (perception) and selecting actions that shape future observations (action). This process naturally balances exploration (uncertainty reduction) and exploitation (goal achievement).

Key processes in Active Inference include:
\begin{itemize}
    \item \textbf{Perception:} Updating beliefs about the hidden states of the environment using variational inference to align the approximate posterior \( Q(x) \) with the true posterior \( P(x|y) \).
    \item \textbf{Action:} Selecting actions that minimize free energy, shaping future observations to conform to the agent’s generative model.
\end{itemize}

Unlike traditional reinforcement learning, which separates reward maximization and exploration into distinct objectives, Active Inference unifies these objectives by incorporating uncertainty reduction as an intrinsic component of decision-making. This intrinsic motivation encourages agents to explore uncertain states while also achieving extrinsic goals.

\subsection{Monte Carlo Tree Search (MCTS)}

Monte Carlo Tree Search (MCTS) is a heuristic search algorithm designed for decision-making in large and complex state spaces \citep{browne2012survey}. It incrementally builds a search tree by iteratively simulating playouts, balancing exploration and exploitation to identify promising actions. Each node in the tree represents a state, and edges represent actions.

The MCTS process is typically divided into four key phases:

\begin{itemize}
    \item \textbf{Selection:} Starting from the root node, the algorithm traverses the tree by selecting child nodes based on a specific policy until a leaf node is reached. The most commonly used selection policy is based on Upper Confidence Bounds (UCB), described below.
    
    \item \textbf{Expansion:} If the leaf node does not represent a terminal state, one or more child nodes are added to the tree, representing unexplored actions.
    
    \item \textbf{Simulation:} A simulation, or playout, is performed from the expanded node, where actions are sampled according to a default policy (e.g., random actions) until a terminal state is reached. The return from this simulation provides an estimate of the value of the expanded node.
    
    \item \textbf{Backpropagation:} The results of the simulation are propagated back up the tree, updating the values and visit counts of all nodes along the path.
\end{itemize}

By iteratively performing these steps, MCTS progressively refines its estimates of action values, focusing computational resources on the most promising parts of the search space. This property makes MCTS highly effective in problems with large state spaces and uncertain outcomes.

\subsubsection{Upper Confidence Bound 1 (UCB1)}

The Upper Confidence Bound 1 (UCB1) algorithm is a widely used technique in MCTS to address the exploration-exploitation tradeoff during tree traversal \citep{auer2002finite}. UCB1 assigns a score to each child node, balancing the average reward observed (exploitation) with the uncertainty of the node (exploration).

The UCB1 value for selecting a child node \( i \) is computed as:

\begin{equation}
\label{eq:ucb1}
\text{UCB1}_i = Q_i + C_{\text{ucb}} \sqrt{\frac{\ln N}{N_i}},
\end{equation}

where:

\begin{itemize}
  \item \( Q_i \) is the average reward (mean value) of child \( i \).
  \item \( N \) is the total number of times the parent node has been visited.
  \item \( N_i \) is the number of times child \( i \) has been visited.
  \item \( C_{\text{ucb}} \) is the exploration constant that controls the degree of exploration.
\end{itemize}

The first term, \( Q_i \), promotes exploitation by preferring actions that have yielded higher rewards on average. The second term, \( C_{\text{ucb}} \sqrt{\frac{\ln N}{N_i}} \), encourages exploration by assigning a higher bonus to actions that have been selected fewer times, thus having higher uncertainty. The logarithmic factor \( \ln N \) ensures that as the number of visits \( N \) to the parent node increases, the exploration bonus decreases, allowing the algorithm to focus more on exploitation over time.

By integrating UCB1 into MCTS, the algorithm effectively balances the need to explore new actions that might lead to better rewards with the need to exploit actions that have already shown promising results. This balance is crucial for converging towards optimal policies in decision-making tasks.

\section{Proposed Planner}
\label{sec:proposed_planner}

\begin{figure}[H]
    \centering
    \includegraphics[width=\linewidth]{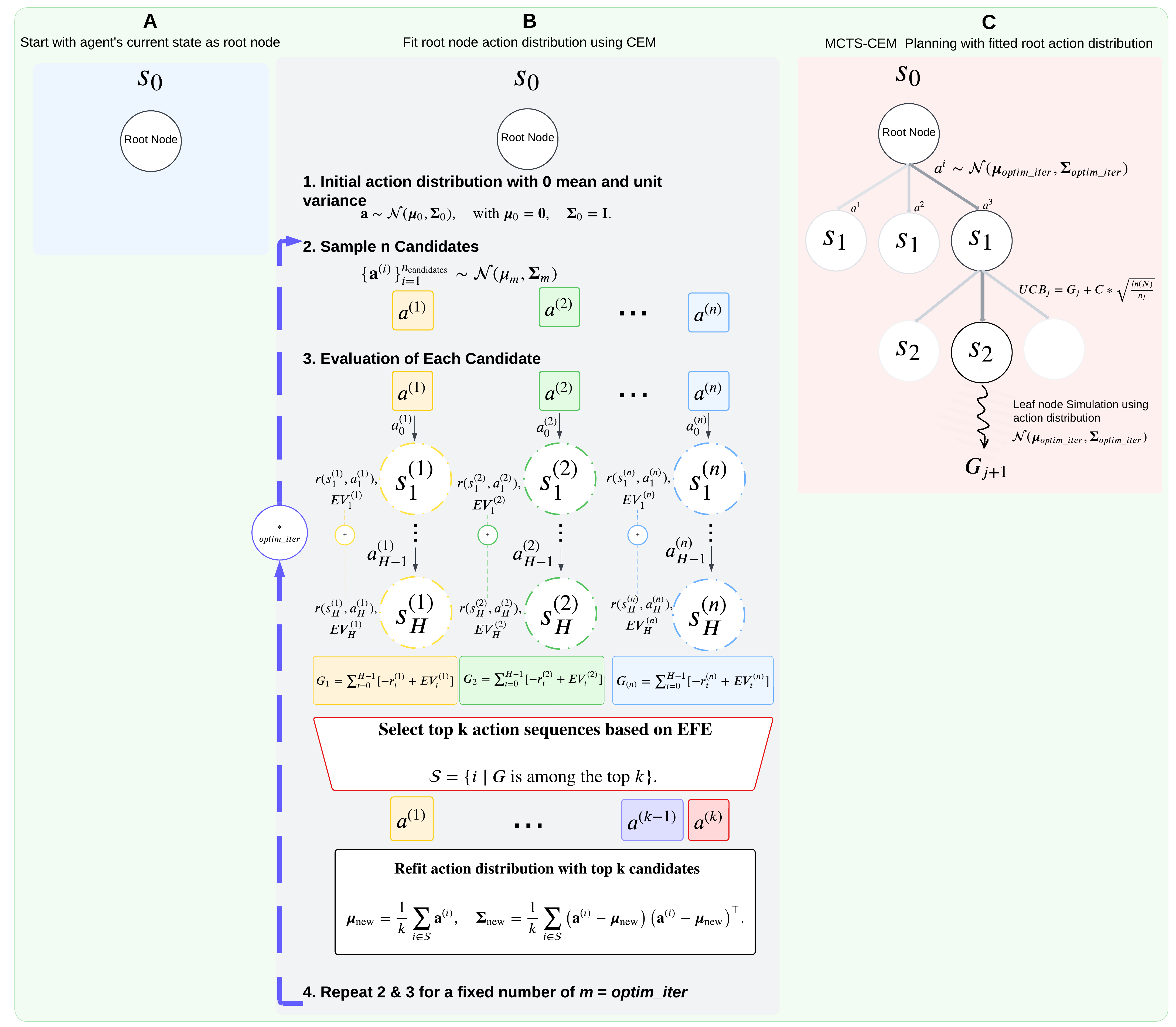}
    \caption{\textbf{MCTS-CEM Diagram.} \\
\textbf{A:} Initialize the MCTS tree with the agent's current state \( s_0 \). \\
\textbf{B:} Fit the root node's action distribution using CEM. Actions are evaluated by minimizing expected free energy (\( G_i \)), with next states sampled using the current action Gaussian. The epistemic value \(  EV^i_t \) is computed as the KL divergence, and rewards (\( r_t^i \)) approximate \( \ln P(y_t^i) \). The top-performing actions refine the distribution iteratively. \\
\textbf{C:} Use the fitted action distribution for action sampling during MCTS exploration, balancing exploitation and exploration with UCB-like selection and consistent simulations at the leaves.}
    \label{fig:mcts-cem}
\end{figure}

We now present our \emph{MCTS-CEM} planning framework, which integrates Monte Carlo Tree Search with the Cross-Entropy Method (CEM) at the \emph{root node} to handle continuous actions effectively. Figure~\ref{fig:mcts-cem} provides a high-level overview, highlighting three main components:
\begin{enumerate}
    \item \textbf{(A)} The root node, initialized with the agent's current state $s_0$.
    \item \textbf{(B)} The process of fitting a \emph{single Gaussian action distribution} using CEM at the root node.
    \item \textbf{(C)} The subsequent tree-based planning (MCTS) stage, which uses the fitted root action distribution for exploration, rollouts, and leaf-node simulations.
\end{enumerate}

\subsection{Root Action Distribution Planning}
\label{sec:root_action_distribution}

The key idea is to learn one Gaussian distribution over actions at the root node, then reuse it throughout tree-based planning and simulation, thereby ensuring consistent estimates of value and reward. We denote this root action distribution by $a \,\sim\, \mathcal{N}\bigl(\boldsymbol{\mu},\,\boldsymbol{\Sigma}\bigr)$.

where $\boldsymbol{\mu}$ and $\boldsymbol{\Sigma}$ are optimized via CEM to maximize expected returns plus any epistemic (information gain) terms.

\subsubsection{Fitting the Root Action Distribution via CEM}
\label{sec:fitting_cem}

\begin{figure}[H]
    \centering
    \includegraphics[width=0.9\linewidth]{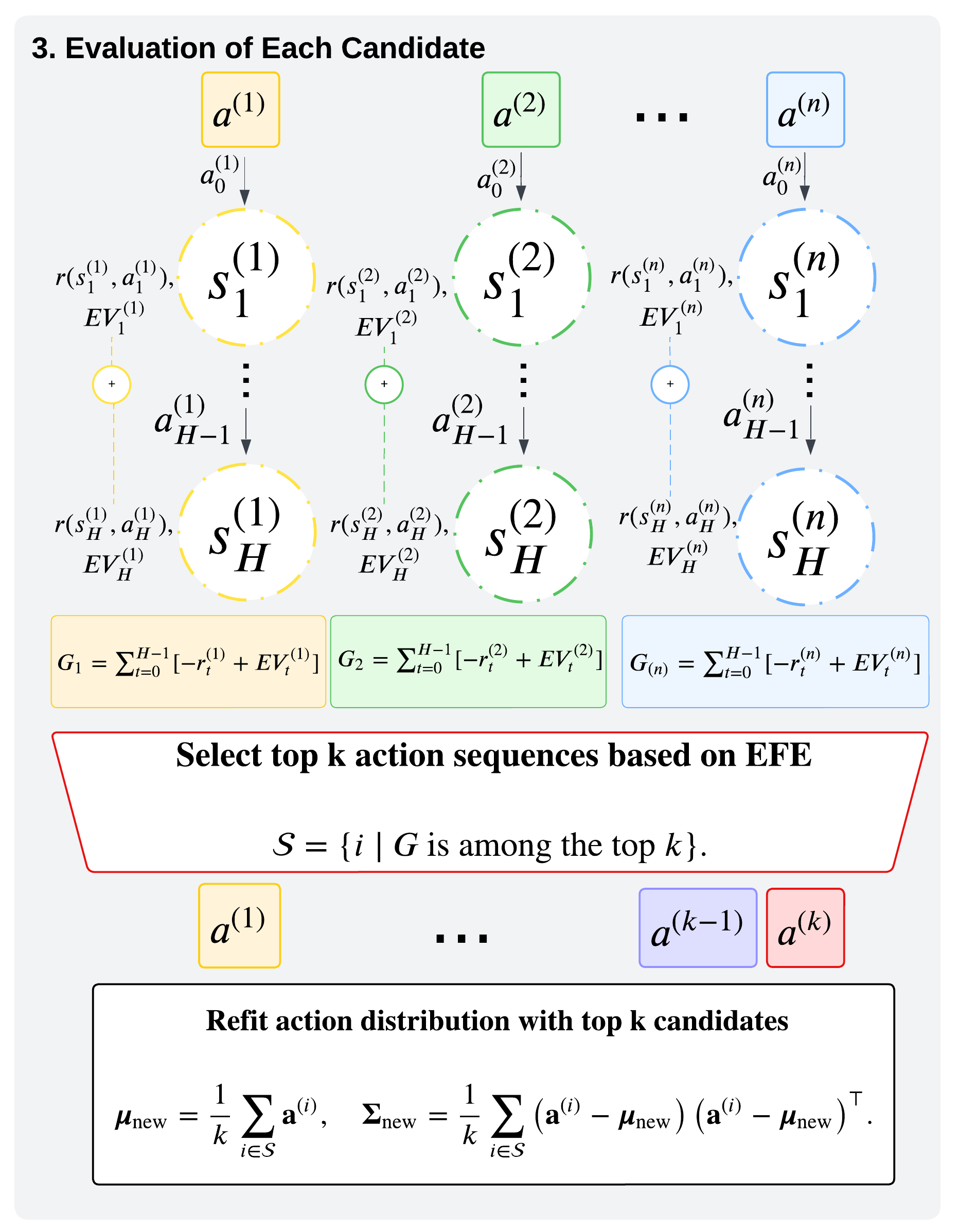}
    \caption{\textbf{Root Action Distribution Fitting Using CEM:} This diagram focuses on subsection \textbf{3} of component \textbf{B} in the MCTS-CEM process. Candidate actions \( \{a^{(1)}, \ldots, a^{(n_{\text{candidates}})} \} \) are sampled from a Gaussian distribution. Their evaluations, based on the expected free energy objective \( G_i \), approximate extrinsic value \(  ln P(y_t^i) \approx r^i_t \) and epistemic value \(  EV^i_t \) from KL divergence. The top \( k \) candidates refine the distribution, optimizing exploration and exploitation.
    }\label{fig:MCTS-CEM-B3}
\end{figure}

At the beginning of planning (from root state $s_0$), we perform Cross-Entropy Method optimization to obtain $\boldsymbol{\mu}$ and $\boldsymbol{\Sigma}$. 
Figure~\ref{fig:MCTS-CEM-B3} illustrates this process (labeled ``3. Evaluation of Each Candidate''):

\begin{enumerate}
    \item \textbf{Initialize} a Gaussian action distribution with mean $\boldsymbol{\mu}_0 = \mathbf{0}$ and covariance $\boldsymbol{\Sigma}_0 = \mathbf{I}$.
    \item \textbf{Sample candidates} $\{\mathbf{a}^{(i)}\}_{i=1}^{n_{\text{candidates}}}$ from the current Gaussian. Each $\mathbf{a}^{(i)}$ is an $H$-step action sequence for the planning horizon.
    \item \textbf{Evaluate each candidate} via short model-based rollouts, scoring it by the sum of extrinsic reward and the epistemic value (an information-gain term) across the horizon.
    \item \textbf{Refit the distribution} to the top-$k$ performers: 

    \begin{equation}
    \label{eq:refit_distribution}
    \boldsymbol{\mu}_{\mathrm{new}} \;=\; \tfrac{1}{k}\!\sum_{i \in S}\mathbf{a}^{(i)}, 
        \quad
        \boldsymbol{\Sigma}_{\mathrm{new}} \;=\; \tfrac{1}{k}\!\sum_{i \in S}(\mathbf{a}^{(i)} - \boldsymbol{\mu}_{\mathrm{new}})(\mathbf{a}^{(i)} - \boldsymbol{\mu}_{\mathrm{new}})^\top,
    \end{equation}
    where $S$ is the index set of top-$k$ candidates.
    \item \textbf{Repeat} for a fixed number of CEM iterations until convergence.
\end{enumerate}

The output is a single, optimized action distribution 
\(
\mathcal{N}(\boldsymbol{\mu}, \boldsymbol{\Sigma})
\)
centered on promising actions from the model’s perspective.

\subsubsection{MCTS-CEM Planning with the Fitted Root Action Distribution}
\label{sec:mcts_cem_planning}

Once the root action distribution has been fit via CEM, we keep it fixed for the duration of the tree-based planning. Figure~\ref{fig:MCTS-CEM-C} depicts this stage:

\begin{figure}[H]
\centering
\includegraphics[width=0.7\textwidth]{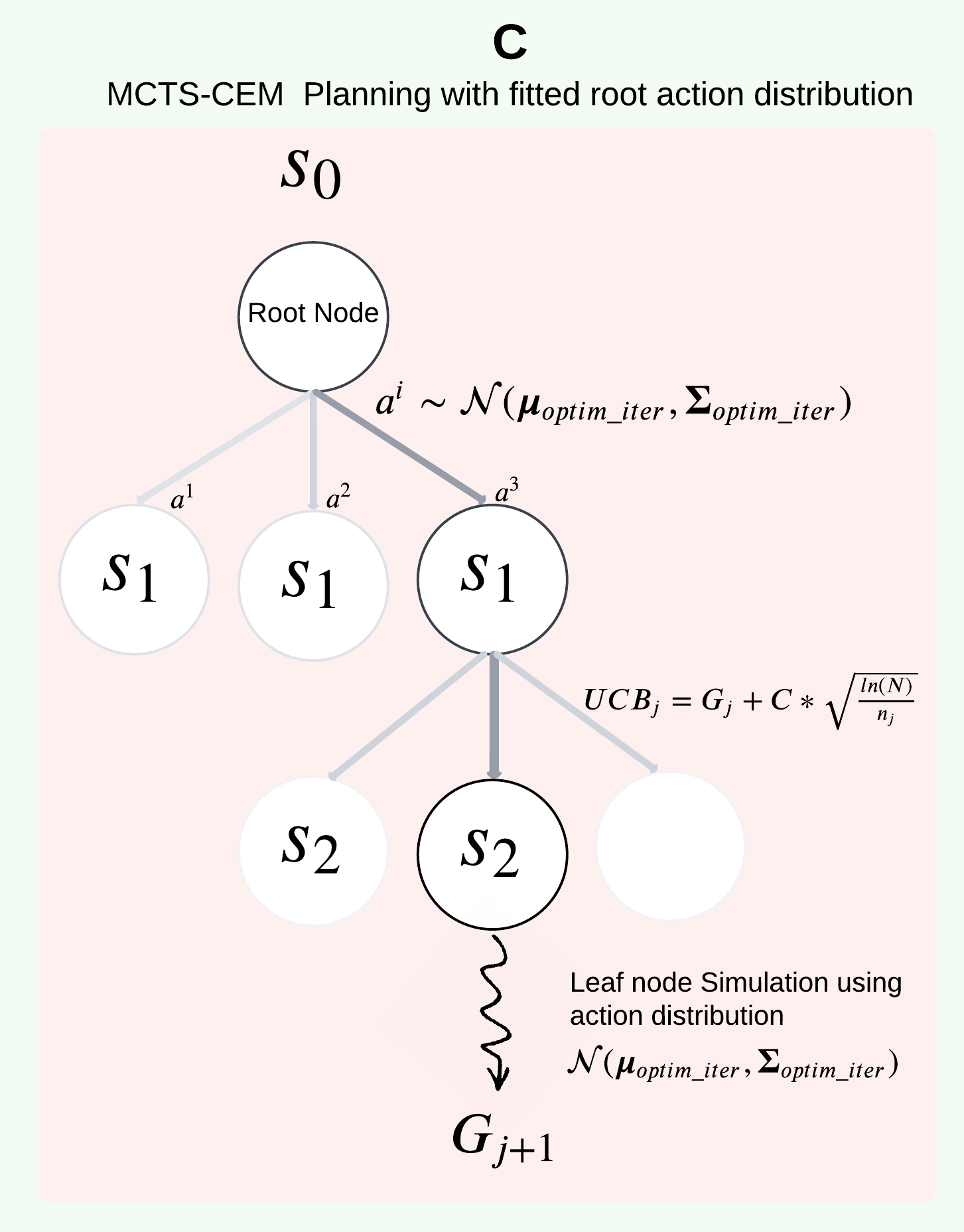}
\caption{%
\textbf{MCTS-CEM Planning (Component C).}
After fitting the root action distribution, MCTS uses it to drive action sampling at each expansion step and for leaf-node simulations.%
}
\label{fig:MCTS-CEM-C}
\end{figure}

\paragraph{MCTS Expansion and Action Selection.}
When selecting actions at each decision node, we sample from 
\(
   \mathcal{N}(\boldsymbol{\mu},\,\boldsymbol{\Sigma})
\)
rather than optimizing new distributions at non-root nodes. This design assumes that states near the root are sufficiently representative of what we will encounter deeper in the tree; hence, a single Gaussian can remain effective as the agent expands its lookahead. If $N_{\text{children}}$ new actions must be considered, we draw $N_{\text{children}}$ distinct samples from the root distribution to populate child nodes.

\paragraph{Simulation/Rollout.}
When a leaf node is reached and we need to estimate its value (i.e., to “roll out” until horizon), we again sample future actions from the same root distribution. This ensures that the value estimates at leaf nodes reflect outcomes under the \emph{same} policy used to expand the tree, resulting in consistent planning. In practice, short rollouts can be performed, or we can simply evaluate a truncated horizon to keep computations manageable. A recognized limitation is that as states become farther from the root, the originally fitted action distribution may become suboptimal for these distant states. This can lead to rollouts that underestimate the true value if the root distribution does not align well with deeper states. Nonetheless, because we use the same distribution throughout, our value estimates remain consistent with the policy we have committed to at the root.

By maintaining one fitted action distribution at the root and using it consistently throughout expansions and rollouts, \emph{MCTS-CEM} ties together the policy used to score states with the policy used to explore them. This avoids needless re-optimization of actions at every node. While this approach may yield suboptimal action choices in states that significantly diverge from the root, it provides self-consistent planning in terms of policy and value estimation. Addressing this limitation to enable broader applicability to larger search horizons is an exciting area for future work. Empirical results in Section~\ref{sec:experiments} demonstrate that this strategy\textemdash even with a truncated horizon\textemdash leads to robust performance gains, especially in sparse or high-dimensional continuous tasks.

\subsection{Epistemic Value as Information Gain Bonus}

The expected free energy \( G(\pi) \) for a policy \( \pi \) can be decomposed into two key components: the extrinsic value (preferences over observations) and the epistemic value (information gain). Formally, it is defined as \citep{friston2015active}:

\begin{equation}
\label{eq:expected_free_energy}
G(\pi) = \mathbb{E}_{q(s_{1:H} | \pi)} \left[ \sum_{t=1}^H \underbrace{ - \ln P(o_t | s_t) }_{\text{Extrinsic Value}} + \underbrace{ D_{\text{KL}} \left[ q(s_t | \pi) \, \| \, p(s_t | s_{t-1}, a_{t-1}) \right] }_{\text{Epistemic Value}} \right],
\end{equation}

where \( q(s_{1:H} | \pi) \) is the variational posterior over states given policy \( \pi \), \( P(o_t | s_t) \) is the likelihood of observations given states, \( D_{\text{KL}}[\cdot \, \| \, \cdot] \) is the Kullback-Leibler (KL) divergence, and \( p(s_t | s_{t-1}, a_{t-1}) \) is the prior predictive distribution of states under the policy. The extrinsic value encodes the agent's preferences over observations, while the epistemic value quantifies the expected information gain about the environment's dynamics.

In reinforcement learning, the agent's goal is to maximize cumulative rewards. To incorporate the Free Energy Principle, we approximate the extrinsic value \( -\ln P(o_t | s_t) \) with the negative reward function \( -r(s_t, a_t) \). This is a standard practice when aligning reinforcement learning with the Free Energy framework, as rewards are viewed as representing the agent's preferences over states or outcomes \citep{friston2009reinforcement, tschantz2020scaling, millidge2020deep}. This approximation allows us to interpret reward maximization as a form of free energy minimization, reframing the agent's extrinsic motivation in terms of the principle.

The epistemic value measures how much an agent's belief about the next state changes when new information is available, capturing the expected information gain from taking an action. For a candidate action sequence \( \mathbf{a}^i \), the epistemic value at time \( t \) is defined as the expected KL divergence between the approximate posterior and prior predictive distributions.

\begin{equation}
\label{eq:epistemic_value_kl}
\text{EV}^i_t = \mathbb{E}_{q(s_{t+1} | s_t^i, a_t^i)} \left[ \ln q(s_{t+1} | s_t^i, a_t^i) - \ln p(s_{t+1} | s_t^i, a_t^i) \right].
\end{equation}

To compute the epistemic value practically, we approximate the expected KL divergence as the difference between the entropy of the aggregated predictive distribution and the average entropy of the individual model predictions. This approximation is inspired by the Bayesian Active Learning by Disagreement (BALD) framework \citep{houlsby2011bayesian, gal2017deep}, where mutual information is used to quantify epistemic uncertainty.

Starting from Equation~\eqref{eq:epistemic_value_kl} and with some minor abuse of notation, we observe that the log-ratio 
\(\ln q(s_{t+1} \mid s_t^i, a_t^i) - \ln p(s_{t+1} \mid s_t^i, a_t^i) = ln\frac{q(s)}{p(s)}\). Taking the expectation under the posterior \(q(\cdot)\), we have

\begin{align}
\label{eq:kl_divergence}
D_{\text{KL}}[ q(s) \| p(s) ] &= \int q(s) \ln \frac{q(s)}{p(s)} \, ds \\
&= \int q(s) \ln q(s) \, ds - \int q(s) \ln p(s) \, ds \\
&= -H(q) - \int q(s) \ln p(s) \, ds,
\end{align}

where \( H(q) = -\int q(s) \ln q(s) \, ds \) is the entropy of \( q(s) \). The term \( \int q(s) \ln p(s) \, ds \) can be challenging to compute directly. However, if \( q(s) \) and \( p(s) \) are both Gaussian distributions or mixtures of Gaussians, and they are close to each other (i.e., \( q(s) \approx p(s) \)), we can approximate this term.

Since \( q(s) \) is the aggregated predictive distribution from the ensemble, and \( p(s) \) represents the individual model predictions, we can approximate:

\begin{equation}
\label{eq:approx_log_p}
\int q(s) \ln p(s) \, ds \approx \frac{1}{M} \sum_{m=1}^M \int q(s) \ln p_m(s) \, ds.
\end{equation}

The cross-entropy between \( q(s) \) and \( p(s) \) is defined as:

\begin{equation}
\label{eq:cross_entropy}
H(q, p) = -\int q(s) \ln p(s) \, ds.
\end{equation}

Therefore, we have:

\begin{equation}
\label{eq:expected_log_p}
\int q(s) \ln p(s) \, ds = -H(q, p).
\end{equation}

Assuming that \( q(s) \) is close to \( p(s) \), the cross-entropy \( H(q, p) \) can be approximated by the entropy of \( p(s) \):

\begin{equation}
\label{eq:approx_cross_entropy}
H(q, p) \approx H(p).
\end{equation}

Similarly, for each ensemble model \( p_m(s) \):

\begin{equation}
\label{eq:approx_cross_entropy_m}
\int q(s) \ln p_m(s) \, ds = -H(q, p_m) \approx -H(p_m).
\end{equation}

Substituting back into the expression for the KL divergence, we obtain:

\begin{align}
\label{eq:kl_divergence_approx}
D_{\text{KL}}[ q \| p ] &\approx -H(q) - \left( -H(p) \right) = -H(q) + H(p).
\end{align}

Similarly, using Equation \eqref{eq:approx_log_p}, we have:

\begin{align}
\label{eq:kl_divergence_approx_m}
D_{\text{KL}}[ q \| p ] &\approx -H(q) + \frac{1}{M} \sum_{m=1}^M H(p_m).
\end{align}

Therefore, the epistemic value becomes:

\begin{align}
\label{eq:epistemic_value_entropy_difference}
\text{EV}_{t,i} &= D_{\text{KL}}[ q(s_{t+1} | s_t^i, a_t^i) \| p(s_{t+1} | s_t^i, a_t^i) ] \\
&\approx -H\left( q(s_{t+1} | s_t^i, a_t^i) \right) + \frac{1}{M} \sum_{m=1}^M H\left( p_m(s_{t+1} | s_t^i, a_t^i) \right).
\end{align}

Since \( q(s_{t+1} | s_t^i, a_t^i) \) is the aggregated predictive distribution \( p(s_{t+1} | s_t^i, a_t^i) \), we can write:

\begin{equation}
\label{eq:final_entropy_difference}
\text{EV}_{t,i} \approx H\left( p(s_{t+1} | s_t^i, a_t^i) \right) - \frac{1}{M} \sum_{m=1}^M H\left( p_m(s_{t+1} | s_t^i, a_t^i) \right).
\end{equation}

This approximation relies on the assumptions that the individual model distributions from the ensemble \( p_m(s_{t+1} | s_t^i, a_t^i) \) and the aggregated distribution \( p(s_{t+1} | s_t^i, a_t^i) \) are approximately Gaussian, and that the ensemble disagreement reflects the epistemic uncertainty about the environment's dynamics.

In the BALD framework \citep{houlsby2011bayesian, gal2017deep}, the mutual information between predictions and model parameters is expressed as:

\begin{equation}
\label{eq:mutual_information}
I[y; \theta | x] = H\left( \mathbb{E}_{p(\theta)}[p(y | x, \theta)] \right) - \mathbb{E}_{p(\theta)}\left[ H\left( p(y | x, \theta) \right) \right],
\end{equation}

where \( y \) is the prediction, \( x \) is the input, and \( \theta \) represents the model parameters. This parallels our expression for the epistemic value, where the first term is the entropy of the aggregated predictions, and the second term is the average entropy over the models.

Calculating the entropy of the aggregated predictive distribution \( H\left( p(s_{t+1} | s_t^i, a_t^i) \right) \) directly is challenging since it is a mixture of Gaussians without a closed-form entropy expression. To estimate this entropy, we employ the Kozachenko–Leonenko entropy estimator \citep{kozachenko1987sample}, a non-parametric method based on \( k \)-nearest neighbor distances among samples. The Kozachenko–Leonenko estimator for a set of \( N \) samples \( \{x_i\}_{i=1}^N \) in \( \mathbb{R}^d \) is given by:

\begin{equation}
\label{eq:kl_entropy_estimator}
H_{\text{Kozachenko-Leonenko}} \approx \psi(N) - \psi(k) + \ln c_d + \frac{d}{N} \sum_{i=1}^N \ln \epsilon_i,
\end{equation}

where \( \psi(\cdot) \) is the digamma function, \( c_d \) is the volume of the \( d \)-dimensional unit ball, \( \epsilon_i \) is twice the distance from sample \( x_i \) to its \( k \)-th nearest neighbor, and \( d \) is the dimensionality of the state space. In our implementation, we generate samples from the aggregated predictive distribution by sampling from each ensemble model's output and aggregating the results. We set \( k = 3 \) for the \( k \)-nearest neighbors and compute pairwise distances between samples to find \( \epsilon_i \). Using Equation \eqref{eq:kl_entropy_estimator}, we estimate \( H\left( p(s_{t+1} | s_t^i, a_t^i) \right) \).

For the individual model entropies \( H\left( p_m(s_{t+1} | s_t^i, a_t^i) \right) \), we use the closed-form expression for the entropy of a Gaussian distribution:

\begin{equation}
\label{eq:gaussian_entropy}
H(\mathcal{N}(\mu, \Sigma)) = \frac{1}{2} \ln \left[ (2\pi e)^d \det \Sigma \right],
\end{equation}

where \( \Sigma \) is the covariance matrix (assumed diagonal in our case).

Our planner aims to minimize the expected free energy by balancing the trade-off between exploiting known rewarding actions and exploring uncertain regions. We define the combined objective function for candidate \( i \) as:

\begin{equation}
\label{eq:combined_return_free_energy}
G_i = \sum_{t=0}^{H-1} \left[ - r(s_t^i, a_t^i) + \lambda \cdot \text{EV}^i_t \right],
\end{equation}

where \( r(s_t^i, a_t^i) \) is the reward function, \( \text{EV}^i_t \) is the epistemic value (Information Gain) at time \( t \), and \( \lambda \) is a weighting factor balancing exploitation and exploration.

During planning, we generate a set of candidate action sequences \( \{ \mathbf{a}_i \} \) and evaluate them using the combined objective function \( G_i \). We use the Cross-Entropy Method (CEM) to iteratively refine the action distribution by selecting the top-performing candidates and fitting a Gaussian distribution over them. The process involves initializing the action distribution's mean and standard deviation, sampling candidate action sequences, evaluating them using Equation \eqref{eq:combined_return_free_energy}, selecting the top performers, updating the action distribution based on these candidates, and repeating the process for a fixed number of iterations. By minimizing \( G_i \), the planner selects actions that are expected to yield high rewards while also providing valuable information about the environment.

Our approach ensures a principled balance between exploration and exploitation. The negative reward term \( - r(s_t^i, a_t^i) \) encourages the agent to select actions that maximize expected rewards, aligning with the goal of reward maximization in reinforcement learning. The epistemic value \( \lambda \cdot \text{EV}^i_t \) incentivizes the agent to choose actions that reduce uncertainty, leading to better knowledge of the environment's dynamics. This balance is crucial for efficient learning, as it prevents the agent from overly exploiting known rewarding actions without improving its understanding of the environment.

\subsection{Algorithmic Details and Pseudocode}

Below, we provide the pseudocode for MCTS-CEM (Algorithm~\ref{alg:mcts_cem_root}), which illustrates how the root action distribution is used during planning. This algorithm reflects the approach described in Sections~\ref{sec:root_action_distribution}--\ref{sec:mcts_cem_planning}, including how reward and epistemic value are integrated during the rollouts.

\begin{algorithm}[H]
\scriptsize 
\renewcommand{\thealgorithm}{1}
\renewcommand{\algorithmicensure}{\textbf{Output:}}
\renewcommand{\algorithmicrequire}{\textbf{Input:}}
\caption{MCTS-CEM with Root Action Distribution}
\label{alg:mcts_cem_root}
\begin{algorithmic}[1]
\REQUIRE State $s_0$, fitted Gaussian $\mathcal{N}(\boldsymbol{\mu},\boldsymbol{\Sigma})$, ensemble $\{p_m\}$, 
         reward $r(\cdot,\cdot)$, horizon $H$, \# sims $N_{\mathrm{sim}}$, $c$, $\gamma$
\ENSURE Selected action $a^*$

\STATE Create root node $v_0$ with state $s_0$.
\FOR{$\mathrm{sim}=1 \ldots N_{\mathrm{sim}}$}
   \STATE \textbf{Selection:} \quad $v \gets v_0$
   \WHILE{$v$ not terminal}
       \IF{$v$ not fully expanded}
           \STATE \textbf{break}
       \ELSE
           \STATE Select child $v'$ via UCB: 
           \[
             v' = \arg\max_{c \in \mathrm{Children}(v)}\![ Q(c) + c \sqrt{\tfrac{\ln N_v}{N_c}} ]
           \]
           \STATE $v \gets v'$
       \ENDIF
   \ENDWHILE
   \STATE \textbf{Expansion:}
   \IF{$v$ not fully expanded \textbf{and} not terminal}
       \STATE Sample $\mathbf{a} \sim \mathcal{N}(\boldsymbol{\mu},\boldsymbol{\Sigma})$
       \STATE $s_{\mathrm{new}} \gets f(s_v,\mathbf{a})$, create child node $v_{\mathrm{new}}$
       \STATE Add $v_{\mathrm{new}}$ to Children($v$); $v \gets v_{\mathrm{new}}$
   \ENDIF
   \STATE \textbf{Simulation (Rollout):} \quad $G \gets 0,\; s \gets s_v$
   \FOR{$t=0 \ldots \mathrm{rolloutH}-1$}
       \STATE $a \sim \mathcal{N}(\boldsymbol{\mu},\boldsymbol{\Sigma})$
       \STATE $r_t \gets r(s,a)$;\; $G \gets G + \gamma^t r_t$
       \STATE $s \gets \mathrm{mean}\{p_m(s \mid s,a)\}$ \COMMENT{(or sample)}
   \ENDFOR
   \STATE \textbf{Backpropagation:} \quad Propagate $G$ up the tree
\ENDFOR
\STATE \textbf{Action Selection:} 
\STATE $a^* \gets \arg\max_{a_c} N_{c}\,\text{(children of }v_0)$
\RETURN $a^*$
\end{algorithmic}
\end{algorithm}

\section{Experiments}
\label{sec:experiments}

In this section, we present experiments conducted to evaluate our proposed MCTS-CEM. We compare MCTS-CEM to two other planners:

\begin{enumerate}
    \item \textbf{CEM Planner}: A regular Cross Entropy Method planner that uses the same model-based rollout used by \cite{tschantz2020reinforcement} during the simulate phase of our MCTS-CEM algorithm.
    \item \textbf{MCTS-Random}: An MCTS planner that employs a random policy during rollouts. Critically, the simulate phase of MCTS-Random does not use any Free Energy Minimization for action selection. Instead, at each step of the planning horizon, it samples a random action from a uniform distribution bounded by the environment's action space.
\end{enumerate}

We compare these three planners across five different environments. Specifically, in the \textit{Pendulum} and \textit{Sparse Mountain Car} environments, we run each planner for 10 episodes per trial, while in \textit{HalfCheetah-Run} and \textit{HalfCheetah-Flip} we run each for 1000 episodes. We repeat every trial five times with different random seeds to account for variability in performance. All planners are configured with the same planning horizon and number of simulations per planning step within each environment to ensure a fair comparison. The experiments use the same model dynamics and reward function across all planners.

\subsection{Pendulum}

\begin{figure}[H]
    \centering
    \includegraphics[width=0.3\linewidth]{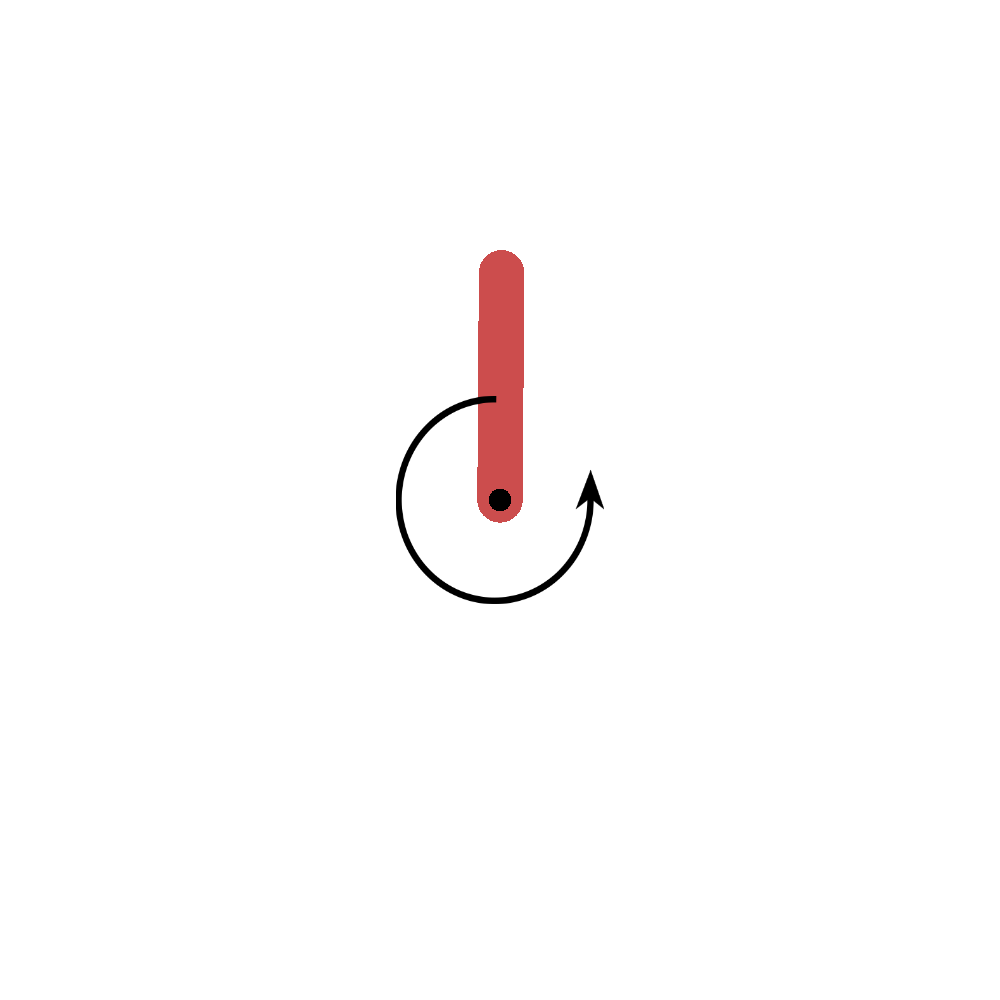}
    \caption{The Pendulum environment, where the agent applies torque to swing the pendulum to its upright position.}
    \label{fig:pendulum_env}
\end{figure}

\begin{figure}[H]
    \centering
    \includegraphics[width=1.2\linewidth]{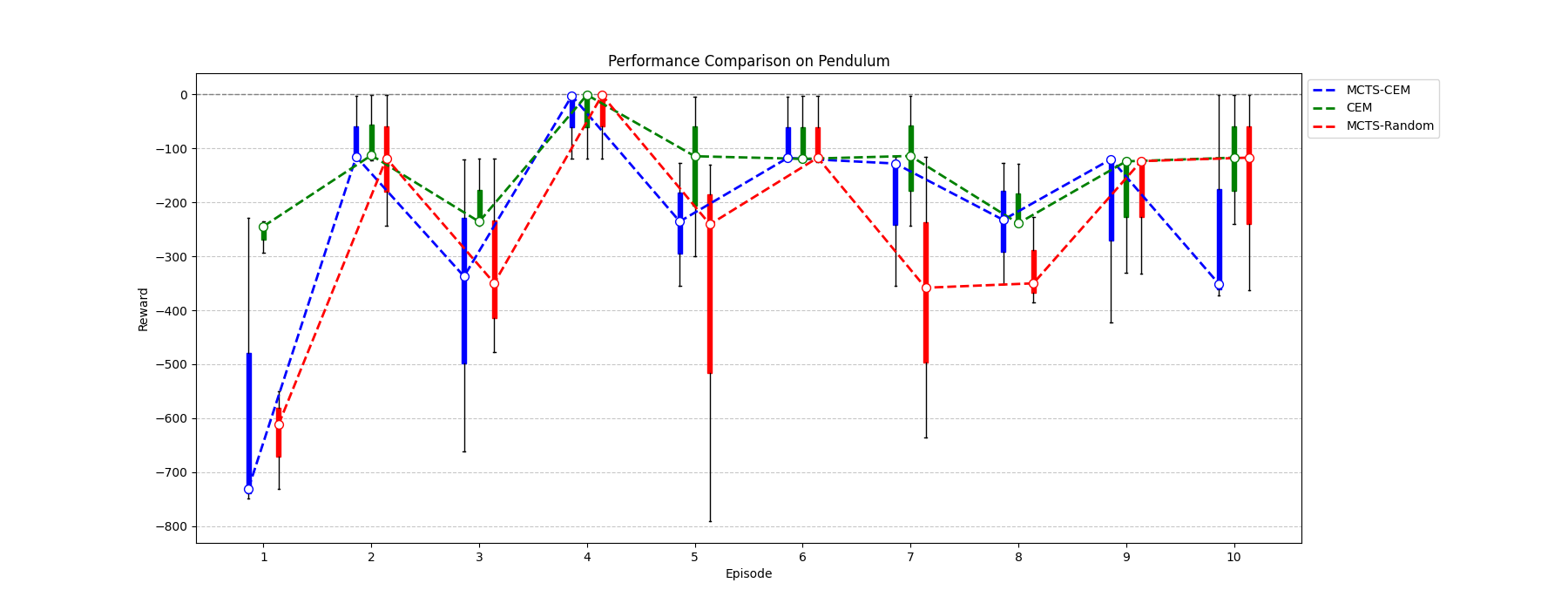}
    \caption{
        Performance comparison of MCTS-CEM, CEM, and MCTS-Random on the Pendulum environment, showing the cumulative reward over episodes averaged over five trials with different random seeds. Error bars represent the standard deviation across trials. In this well-shaped, deterministic setting, the additional exploration provided by MCTS-CEM results in performance comparable to CEM, indicating limited added benefit beyond a straightforward Cross-Entropy Method approach.
    }
    \label{fig:pendulum}
\end{figure}

In the Pendulum environment (Figure~\ref{fig:pendulum_env}), we observe that while MCTS-CEM consistently outperforms MCTS-Random, it does not provide a significant improvement over CEM. This can be attributed to the nature of the environment, which features a one-dimensional action space representing the torque applied to the pendulum’s free end. The reward function is well-shaped and continuous:

\begin{equation}
\label{eq:pendulum_reward_function}
r = -(\theta^2 + 0.1 \cdot \dot{\theta}^2 + 0.001 \cdot \text{torque}^2),
\end{equation}

where \(\theta\) is the pendulum’s angle, normalized between \([-\pi, \pi]\), with 0 being the upright position. The minimum reward is \(-16.27\) when the pendulum is fully displaced with maximum velocity, and the maximum reward is 0 when the pendulum is perfectly upright with no torque applied. Given the simple nature of this reward structure and the deterministic dynamics of the environment, even random action selection by MCTS-Random can achieve maximum reward in the early episodes.

The comparable performance between CEM and MCTS-CEM suggests that the additional exploration performed by MCTS-CEM might be unnecessary in this well-defined and deterministic environment. Once the agent achieves the maximum reward state, further exploration does not add value, as MCTS-CEM continues to search novel state-action pairs even when the optimal policy is already known. The exploration term in MCTS encourages novelty, which, while beneficial in more complex environments, may be counterproductive here where a known deterministic policy is sufficient to consistently achieve optimal performance. Our findings align with those of \citet{bellemare2016unifying}, who demonstrated that in environments with sparse rewards, methods incorporating intrinsic motivation—such as count-based exploration strategies—are particularly effective.  In contrast, in simpler, well-shaped environments, relying on established optimal policies, as CEM does, might be more efficient than the continual planning that MCTS-CEM performs. 

\subsection{Sparse Mountain Car}

\begin{figure}[H]
    \centering
    \includegraphics[width=0.3\linewidth]{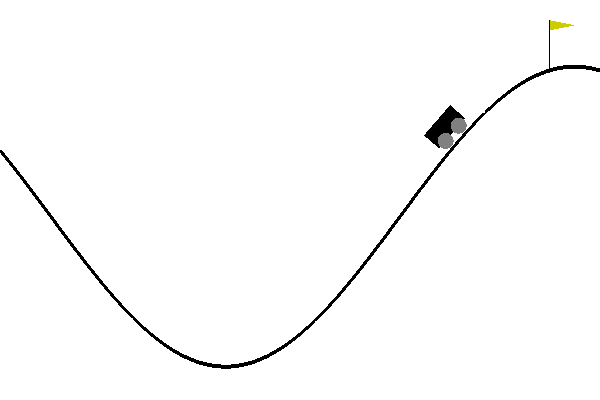}
    \caption{The Sparse Mountain Car environment, where the agent must navigate a car to the flag by building momentum.}
    \label{fig:mountain_car_env}
\end{figure}

\begin{figure}[H]
    \centering
    \includegraphics[width=1.2\linewidth]{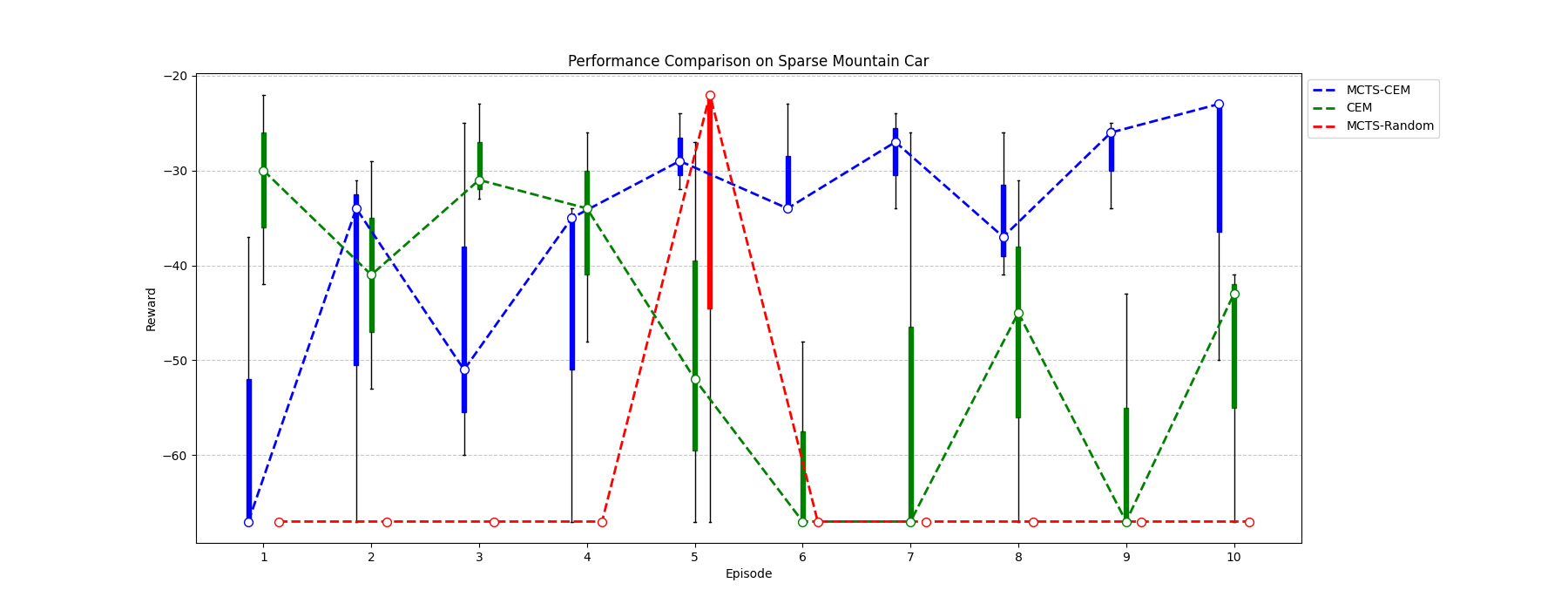}
    \caption{
        Performance comparison of MCTS-CEM, CEM, and MCTS-Random on the Sparse Mountain Car environment, showing the cumulative reward over episodes averaged over five trials with different random seeds. Error bars represent the standard deviation across trials. Although CEM initially outperforms MCTS-CEM due to its reliance on an approximate reward model, MCTS-CEM surpasses both baselines in later episodes by refining reward estimates and leveraging broader UCB-based exploration to uncover high-reward trajectories.
    }
    \label{fig:mountain_car}
\end{figure}

In the Sparse Mountain Car environment (Figure~\ref{fig:mountain_car_env}), the goal is to generate controls that drive a car to the top of a hill marked by a flag, representing the goal state. The car lacks sufficient acceleration to climb the hill directly. Instead, it must first move backward up a smaller hill to gain enough momentum to ascend the larger hill in front. In this sparse reward setting, the agent receives a positive reward of $+1$ only upon reaching the goal state, while incurring a negative reward at every time step it does not reach the goal.

Results on the Sparse Mountain Car environment (Figure~\ref{fig:mountain_car}) highlight the strengths of MCTS-CEM, particularly in later episodes. Initially, CEM outperforms MCTS-CEM, but as more episodes are played, MCTS-CEM improves significantly, ultimately achieving higher maximum rewards than both CEM and MCTS-Random. This improvement can be attributed to the synergy between CEM’s root distribution optimization and the UCB-based tree search within MCTS-CEM. The UCB formula for child selection in our planner is given by:

\begin{equation}
\label{eq:ubc}
\text{UCB}_i = Q_i + c \cdot \sqrt{\frac{\ln N_v}{N_i}},
\end{equation}

where $Q_i$ is the estimated return of child $i$, $N_v$ is the visit count of the parent node $v$, and $N_i$ is the visit count of the child itself. Initially, MCTS-CEM’s double reliance on an approximate reward model can cause performance to lag behind CEM, since reward models may overestimate certain state-action pairs \citep{pathak2017curiosity}. However, as more data is gathered over multiple episodes, the reward model becomes more accurate. Combined with the broader search from UCB expansions, MCTS-CEM discovers high-reward trajectories more effectively than CEM in this sparse environment. 

The ability of MCTS-CEM to tap into a wider range of potential future states with more accurate reward estimates allows it to excel in environments like Mountain Car, where momentum and long-term planning are key to achieving success. This demonstrates the potential of MCTS-CEM in sparse reward environments, where effective planning from multiple future states is crucial for overcoming challenges like sparse rewards and delayed returns \citep{sutton2018reinforcement}.

\subsection{HalfCheetah-Run}

\begin{figure}[H]
    \centering
    \includegraphics[width=0.3\linewidth]{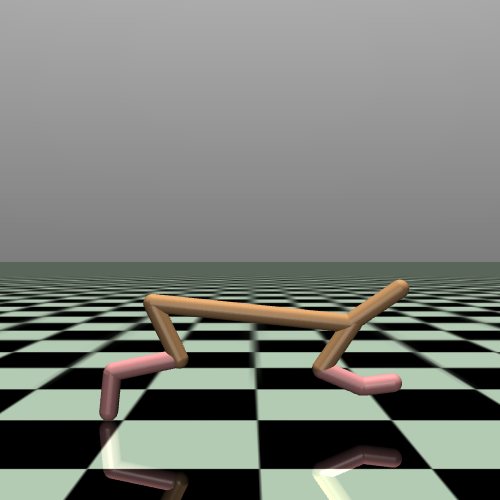}
    \caption{The HalfCheetah-Run environment, where the agent controls a two-dimensional cheetah to maximize forward velocity.}
    \label{fig:halfcheetah_env}
\end{figure}

\begin{figure}[H]
    \centering
    \includegraphics[width=1.2\linewidth]{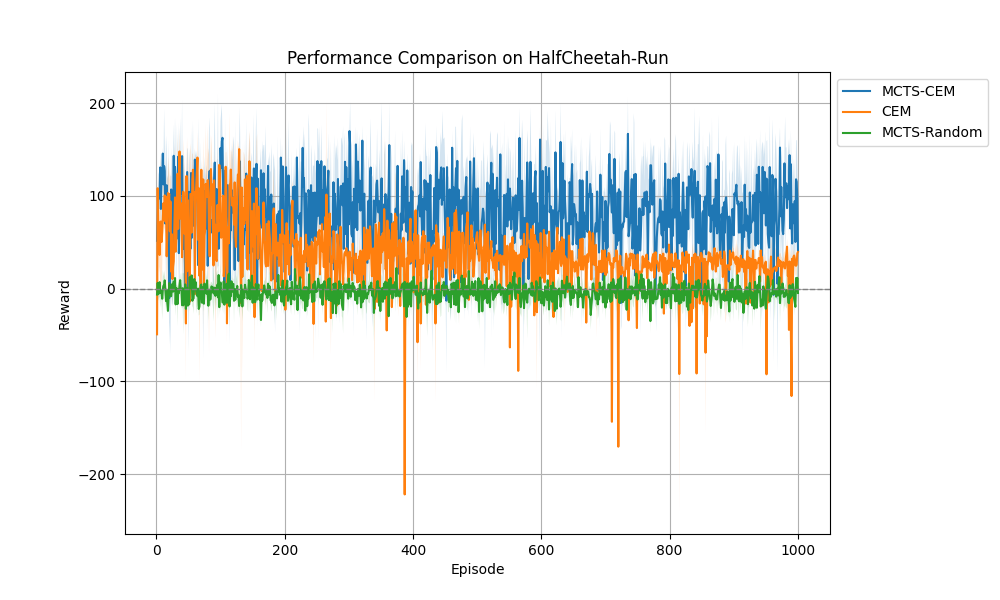}
    \caption{
        Performance comparison of MCTS-CEM, CEM, and MCTS-Random on the HalfCheetah-Run environment. Each curve shows the reward over episodes averaged across five trials with different random seeds. The shaded regions around each curve represent the standard deviation across those trials. MCTS-CEM consistently achieves higher returns by leveraging both UCB-based exploration and CEM’s optimization in continuous action spaces; however, occasional performance dips, or “policy collapse” \citep{millidge2019deep}, \citep{friston2009reinforcement}, highlight the challenge of managing exploration and exploitation in practice.
    }
    \label{fig:halfcheetah_run}
\end{figure}

In the HalfCheetah-Run environment (Figure~\ref{fig:halfcheetah_env}), the objective is for the agent, controlling a simulated two-dimensional cheetah, to run as quickly as possible in the forward direction. The agent receives dense rewards based on its forward velocity, offering frequent feedback that helps MCTS-CEM correct suboptimal trajectories during planning. As illustrated in Figure~\ref{fig:halfcheetah_run}, MCTS-CEM consistently outperforms both the CEM Planner and MCTS-Random by leveraging both upper confidence bound (UCB)-based exploration and CEM’s strength in continuous action optimization.

Despite this strong overall performance, both MCTS-CEM and CEM occasionally exhibit abrupt dips in reward—an effect we refer to as “policy collapse,” following \citet{millidge2019deep} and \citet{friston2009reinforcement}. Here, an imbalance between exploration and exploitation, coupled with potential inaccuracies in the planner’s value or reward estimates, can cause the agent to momentarily overcommit to suboptimal actions. Nevertheless, MCTS-CEM significantly mitigates these collapses compared to plain CEM, likely due to the “double exploration” stemming from both the Monte Carlo tree search (via UCB) and the CEM-based optimization in continuous action space. This dual layer of exploration makes the planner more robust to episodic failures, as it can more quickly identify and correct suboptimal trajectories. In practice, policy collapse can be further alleviated by tuning the weight of intrinsic exploration terms and improving model fidelity (e.g., by  additional training data or larger ensembles), thereby helping to maintain a more reliable balance between exploration and exploitation.

\subsection{HalfCheetah-Flip}

\begin{figure}[H]
    \centering
    \includegraphics[width=0.3\linewidth]{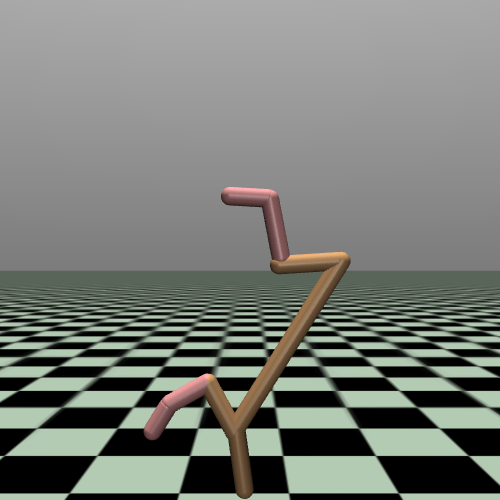}
    \caption{The HalfCheetah-Flip environment, where the agent controls a two-dimensional cheetah to perform front flips.}
    \label{fig:halfcheetah_flip_env}
\end{figure}

\begin{figure}[H]
    \centering
    \includegraphics[width=1.2\linewidth]{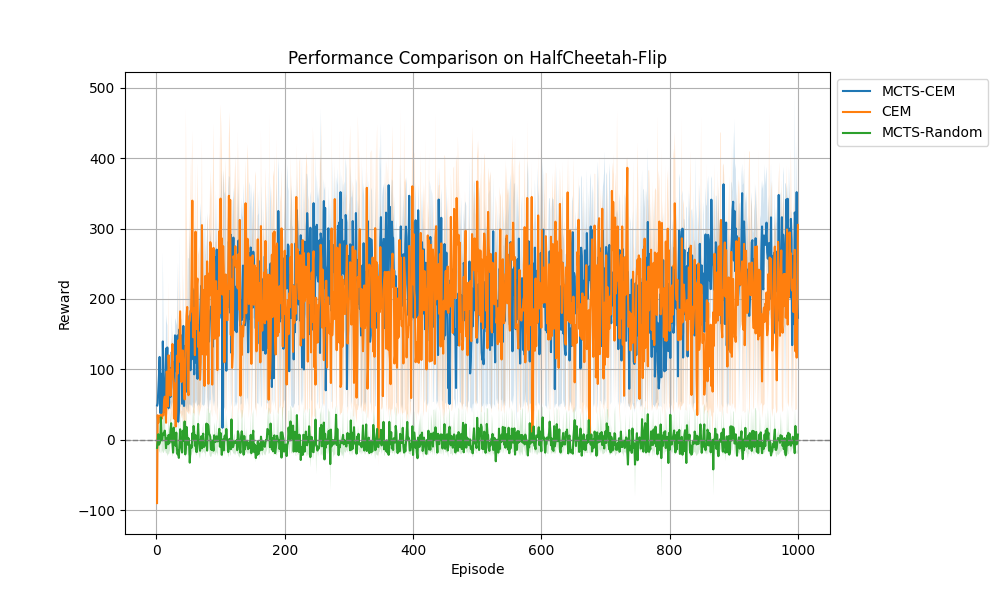}
    \caption{
        Performance comparison of MCTS-CEM, CEM, and MCTS-Random on the HalfCheetah-Flip environment. Each curve shows the reward over episodes averaged across five trials with different random seeds. The shaded regions around each curve represent the standard deviation across those trials. MCTS-CEM and CEM exhibit nearly identical performance, highlighting the relative simplicity of the task for more advanced planners.
    }
    \label{fig:halfcheetah_flip}
\end{figure}

In the HalfCheetah-Flip environment (Figure~\ref{fig:halfcheetah_flip_env}), the objective is for the agent, controlling a simulated two-dimensional cheetah, to perform front flips rather than maximizing forward velocity. The agent receives dense rewards based on the quality and frequency of its flips. Unlike HalfCheetah-Run, the task requires simple motor coordination and involves coarser control over the cheetah's dynamics.

As illustrated in Figure~\ref{fig:halfcheetah_flip}, MCTS-CEM and CEM exhibit nearly identical performance, with no significant advantage observed for the additional planning provided by MCTS-CEM. This is in stark contrast to environments like HalfCheetah-Run, where MCTS-CEM demonstrated a clear performance advantage. The similarity in performance could be attributed to the nature of the task: flipping requires coarser motor control than running, and the additional computational effort of MCTS planning does not yield substantial improvements. Instead, simpler optimization through CEM alone appears sufficient for achieving near-optimal results in this environment.

This result emphasizes the importance of task characteristics in determining the utility of MCTS-CEM. While the algorithm excels in sparse reward environments or well-shaped tasks involving complex, high-dimensional controls, its benefits are less apparent in tasks requiring coarser motor coordination and less intricate action planning. This insight highlights the nuanced trade-offs between computational overhead and planning efficacy in different types of continuous control tasks.

\section{Conclusion}

In this work, we introduced MCTS-CEM, a model-based planner that unifies Monte Carlo Tree Search (MCTS) with the Free Energy Principle’s uncertainty minimization objective. By adopting ensemble-based rollouts and incorporating epistemic value estimates as an intrinsic exploration bonus, our planner naturally balances the drive to maximize extrinsic rewards with the need to reduce uncertainty about the environment’s dynamics. 

A key element of our approach is fitting a single Gaussian action distribution at the root node using the Cross-Entropy Method (CEM). We then use this root distribution consistently throughout the expansion and simulation phases, ensuring that the policy underlying tree search remains coherent with the policy used for value estimation in leaf-node rollouts. This strategy avoids redundant re-optimization at deeper nodes and lends stability to planning. Moreover, the integration of information gain as part of the Free Energy minimization criterion enables a principled exploration mechanism that boosts performance in environments featuring sparse and delayed rewards.

Empirically, we validated MCTS-CEM on a variety of continuous control tasks, including Pendulum, Sparse Mountain Car, and HalfCheetah. Across these benchmarks, our method consistently outperformed or matched baseline planners that either rely solely on CEM or combine MCTS with simplistic (random) policies. Notably, MCTS-CEM demonstrated superior robustness to random seed variability and scaled favorably with increasing amounts of environment interaction, suggesting that its exploration-driven planning is well-suited to long-horizon tasks with limited feedback signals.

Overall, this work highlights the promise of coupling MCTS with Free Energy minimization for active inference in high-dimensional control problems. By merging a powerful search paradigm with an epistemic drive to reduce model uncertainty, we bring a unified view of planning, exploration, and policy refinement to reinforcement learning. However, our results also reveal challenges related to balancing exploration and exploitation, particularly when intrinsic exploration bonuses dominate extrinsic reward optimization, leading to episodic policy collapse. Future research should focus on mitigating this issue by exploring methods such as adaptive regularization of intrinsic exploration bonuses, incorporating uncertainty-aware thresholds, and improving the fidelity of reward models through ensemble methods or additional training. Furthermore, addressing the current limitation of truncated search horizons by extending the planner to deeper or more flexible expansions represents an exciting direction for broadening the applicability of our approach. Addressing these challenges will solidify the application of active inference principles in scalable, model-based reinforcement learning.

\subsection*{Acknowledgments}

This research was conducted as part of the requirements for my PhD. My PhD is supported by funding from Air Force Research Laboratory (AFRL) grant No. FA8650-21-C-1147. Any opinions, findings, conclusions, or recommendations contained herein are those of the authors and do not necessarily represent the official policies or endorsements, either expressed or implied, of the AFRL or the U.S. Government.



\bibliographystyle{APA}

\begin{thebibliography}{100}
\providecommand{\natexlab}[1]{#1}
\expandafter\ifx\csname urlstyle\endcsname\relax
  \providecommand{\doi}[1]{doi:\discretionary{}{}{}#1}\else
  \providecommand{\doi}{doi:\discretionary{}{}{}\begingroup
  \urlstyle{rm}\Url}\fi
  
\bibitem[{Coulom(2006)}]{coulom2006efficient}
Coulom, R. (2006).
\newblock Efficient Selectivity and Backup Operators in Monte-Carlo Tree Search.
\newblock \emph{International Conference on Computers and Games}, \emph{72--83}.

\bibitem[{Browne et~al.(2012)Browne, Powley, Whitehouse, Lucas, Cowling, Rohlfshagen, Tavener, Perez, Samothrakis, \& Colton}]{browne2012survey}
Browne, C.~B., Powley, E., Whitehouse, D., Lucas, S.~M., Cowling, P.~I., Rohlfshagen, P., Tavener, S., Perez, D., Samothrakis, S., \& Colton, S. (2012).
\newblock A survey of Monte Carlo Tree Search methods.
\newblock \emph{IEEE Transactions on Computational Intelligence and AI in Games}, \emph{4}(1), 1--43.

\bibitem[{Houthooft et~al.(2016)Houthooft, Chen, Duan, Schulman, De~Turck, \& Abbeel}]{houthooft2016vime}
Houthooft, R., Chen, X., Duan, Y., Schulman, J., De~Turck, F., \& Abbeel, P. (2016).
\newblock VIME: Variational Information Maximizing Exploration.
\newblock \emph{Advances in Neural Information Processing Systems}, \emph{1109--1117}.

\bibitem[{Lindley(1956)}]{lindley1956measure}
Lindley, D.~V. (1956).
\newblock On a Measure of the Information Provided by an Experiment.
\newblock \emph{The Annals of Mathematical Statistics}, \emph{27}(4), 986--1005.

\bibitem[{Rosin(2011)}]{rosin2011multi}
Rosin, C.~D. (2011).
\newblock Multi-armed bandits with episode context.
\newblock \emph{Annals of Mathematics and Artificial Intelligence}, \emph{61}(3), 203--230.

\bibitem[{Rubinstein(1999)}]{rubinstein1999cross}
Rubinstein, R.~Y. (1999).
\newblock The Cross-Entropy Method for Combinatorial and
Continuous Optimization.
\newblock \emph{Methodology and Computing in Applied Probability}, \emph{1}(2), 127--190.

\bibitem[{Silver et~al.(2016)Silver, Huang, Maddison, Guez, Sifre, van den Driessche, Schrittwieser, Antonoglou, Panneershelvam, Lanctot, et~al.}]{silver2016mastering}
Silver, D., Huang, A., Maddison, C.~J., Guez, A., Sifre, L., van den Driessche, G., Schrittwieser, J., Antonoglou, I., Panneershelvam, V., Lanctot, M., et~al. (2016).
\newblock Mastering the game of Go with deep neural networks and tree search.
\newblock \emph{Nature}, \emph{529}(7587), 484--489.

\bibitem[{Silver et~al.(2017)Silver, Schrittwieser, Simonyan, Antonoglou, Huang, Guez, Hubert, Baker, Lai, Bolton, et~al.}]{silver2017mastering}
Silver, D., Schrittwieser, J., Simonyan, K., Antonoglou, I., Huang, A., Guez, A., Hubert, T., Baker, L., Lai, M., Bolton, A., et~al. (2017).
\newblock Mastering the game of Go without human knowledge.
\newblock \emph{Nature}, \emph{550}(7676), 354--359.

\bibitem[{Cou{\"e}toux et~al.(2011)Cou{\"e}toux, Hoock, Sokolovska, Teytaud \& Bonnard}]{couetoux2011continuous}
Cou{\"e}toux, A., Hoock, J.-B., Sokolovska, N., Teytaud, O., \& Bonnard, N. (2011).
\newblock Continuous Upper Confidence Trees.
\newblock \emph{International Conference on Learning and Intelligent Optimization}, \emph{433--445}.

\bibitem[{Friston(2010)}]{friston2010free}
Friston, K.~J. (2010).
\newblock The free-energy principle: a unified brain theory?
\newblock \emph{Nature Reviews Neuroscience}, \emph{11}(2), 127--138.

\bibitem[{Gelly \& Silver(2007)}]{gelly2007modification}
Gelly, S., \& Silver, D. (2007).
\newblock Combining online and offline knowledge in UCT.
\newblock \emph{Proceedings of the 24th International Conference on Machine Learning}, \emph{273--280}.

\bibitem[{Bubeck et~al.(2011)Bubeck, Munos, Stoltz, \& Szepesvári}]{bubeck2011x}
Bubeck, S., Munos, R., Stoltz, G., \& Szepesvári, C. (2011).
\newblock X-Armed Bandits.
\newblock \emph{Journal of Machine Learning Research}, \emph{12}, 1655--1695.

\bibitem[{Mansley et~al.(2011)Mansley, Weinstein, \& Littman}]{mansley2011sample}
Mansley, C.~R., Weinstein, A., \& Littman, M.~L. (2011).
\newblock Sample-Based Planning for Continuous Action Markov Decision Processes.
\newblock \emph{Proceedings of the 21st International Conference on Automated Planning and Scheduling}, \emph{335--338}.

\bibitem[{Moerland et~al.(2023)Moerland, Broekens, Plaat, \& Jonker}]{moerland2023model}
Moerland, T.~M., Broekens, J., Plaat, A., \& Jonker, C.~M. (2023).
\newblock Model-based reinforcement learning: A survey.
\newblock \emph{Foundations and Trends in Machine Learning}, 16(1), 1--118.


\bibitem[{Chua et~al.(2018)Chua, Calandra, McAllister, \& Levine}]{chua2018deep}
Chua, K., Calandra, R., McAllister, R., \& Levine, S. (2018).
\newblock Deep Reinforcement Learning in a Handful of Trials using Probabilistic Dynamics Models.
\newblock \emph{Advances in Neural Information Processing Systems}, \emph{4754--4765}.

\bibitem[{Hafner et~al.(2019)Hafner, Lillicrap, Norouzi, \& Ba}]{hafner2019learning}
Hafner, D., Lillicrap, T., Norouzi, M., \& Ba, J. (2019).
\newblock Learning Latent Dynamics for Planning from Pixels.
\newblock \emph{Proceedings of the 36th International Conference on Machine Learning}, \emph{2555--2565}.

\bibitem[{Schmidhuber(2010)}]{schmidhuber2010formal}
Schmidhuber, J. (2010).
\newblock Formal Theory of Creativity, Fun, and Intrinsic Motivation (1990–2010).
\newblock \emph{IEEE Transactions on Autonomous Mental Development}, \emph{2}(3), 230--247.

\bibitem[{Nagabandi et~al.(2018)Nagabandi, Kahn, Fearing, \& Levine}]{nagabandi2018neural}
Nagabandi, A., Kahn, G., Fearing, R.~S., \& Levine, S. (2018).
\newblock Neural Network Dynamics for Model-Based Deep Reinforcement Learning with Model-Free Fine-Tuning.
\newblock \emph{IEEE International Conference on Robotics and Automation (ICRA)}, \emph{7559--7566}.

\bibitem[{Lakshminarayanan et~al.(2017)Lakshminarayanan, Pritzel, \& Blundell}]{lakshminarayanan2017simple}
Lakshminarayanan, B., Pritzel, A., \& Blundell, C. (2017).
\newblock Simple and Scalable Predictive Uncertainty
Estimation using Deep Ensembles.
\newblock \emph{Advances in Neural Information Processing Systems}, \emph{6402--6413}.

\bibitem[{De Boer et~al.(2005)De Boer, Kroese, Mannor, \& Rubinstein}]{de2005tutorial}
De Boer, P.-T., Kroese, D.~P., Mannor, S., \& Rubinstein, R.~Y. (2005).
\newblock A Tutorial on the Cross-Entropy Method.
\newblock \emph{Annals of Operations Research}, \emph{134}(1), 19--67.

\bibitem[{Bellemare et~al.(2016)Bellemare, Srinivasan, Ostrovski, Schaul, Saxton, \& Munos}]{bellemare2016unifying}
Bellemare, M.~G., Srinivasan, S., Ostrovski, G., Schaul, T., Saxton, D., \& Munos, R. (2016).
\newblock Unifying Count-Based Exploration and Intrinsic Motivation.
\newblock \emph{Advances in Neural Information Processing Systems}, \emph{1471--1479}.

\bibitem[{Pathak et~al.(2017)Pathak, Agrawal, Efros, \& Darrell}]{pathak2017curiosity}
Pathak, D., Agrawal, P., Efros, A.~A., \& Darrell, T. (2017).
\newblock Curiosity-Driven Exploration by Self-Supervised Prediction.
\newblock \emph{Proceedings of the IEEE Conference on Computer Vision and Pattern Recognition Workshops}, \emph{16--17}.

\bibitem[{Sutton \& Barto(2018)}]{sutton2018reinforcement}
Sutton, R.~S., \& Barto, A.~G. (2018).
\newblock \emph{Reinforcement Learning: An Introduction}.
\newblock MIT Press.


\bibitem[{Guez et~al.(2013)Guez, Silver, \& Dayan}]{guez2013scalable}
Guez, A., Silver, D., \& Dayan, P. (2013).
\newblock Scalable and Efficient Bayes-Adaptive Reinforcement Learning Based on Monte-Carlo Tree Search.
\newblock \emph{Journal of Artificial Intelligence Research}, \emph{48}, 841--883.

\bibitem[{Auer et~al.(2002)Auer, Cesa-Bianchi, \& Fischer}]{auer2002finite}
Auer, P., Cesa-Bianchi, N., \& Fischer, P. (2002).
\newblock Finite-time Analysis of the Multiarmed Bandit Problem.
\newblock \emph{Machine Learning}, \emph{47}(2-3), 235--256.

\bibitem[{Champion et~al.(2022)Champion, Da~Costa, Bowman, \& Grześ}]{champion2022branching}
Champion, T., Da~Costa, L., Bowman, H., \& Grześ, M. (2022).
\newblock Branching Time Active Inference: The theory and its generality.
\newblock \emph{Neural Networks}, \emph{151}, 295--316.

\bibitem[{Fountas et~al.(2020)Fountas, Sajid, Mediano, \& Friston}]{fountas2020deep}
Fountas, Z., Sajid, N., Mediano, P.~A., \& Friston, K. (2020).
\newblock Deep active inference agents using Monte-Carlo methods.
\newblock \emph{Proceedings of the 34th Conference on Neural Information Processing Systems (NeurIPS)}.

\bibitem[{Schwartenbeck et~al.(2018)Schwartenbeck, Passecker, Hauser, FitzGerald, Kronbichler, \& Friston}]{schwartenbeck2018computational}
Schwartenbeck, P., Passecker, J., Hauser, T.~U., FitzGerald, T., Kronbichler, M., \& Friston, K. (2018).
\newblock Computational mechanisms of curiosity and goal-directed exploration.
\newblock \emph{eLife}, \emph{7}, e41703.

\bibitem[{FitzGerald et~al.(2015)FitzGerald, Dolan, \& Friston}]{fitzgerald2015dopamine}
FitzGerald, T.~H.~B., Dolan, R.~J., \& Friston, K. (2015).
\newblock Dopamine, reward learning, and active inference.
\newblock \emph{Frontiers in Computational Neuroscience}, \emph{9}, 136.

\bibitem[{Kocsis and Szepesvári(2006)}]{kocsis2006bandit}
Kocsis, L., \& Szepesvári, C. (2006).
\newblock Bandit Based Monte-Carlo Planning.
\newblock In J. Fürnkranz, T. Scheffer, \& M. Spiliopoulou (Eds.), \emph{Machine Learning: ECML 2006. Lecture Notes in Computer Science} (Vol. 4212, pp. 282--293). Springer, Berlin, Heidelberg.

\bibitem[{Tschantz et~al.(2020)Tschantz, Millidge, Seth, \& Buckley}]{tschantz2020reinforcement}
Tschantz, A., Millidge, B., Seth, A.~K., \& Buckley, C.~L. (2020).
\newblock Reinforcement Learning through Active Inference.
\newblock \emph{Proceedings of the Workshop on Bridging AI and Cognitive Science (ICLR 2020)}.

\bibitem[{Friston et~al.(2015)}]{friston2015active}
Friston, K., Rigoli, F., Ognibene, D., Mathys, C., Fitzgerald, T., \& Pezzulo, G. (2015).
\newblock Active inference and epistemic value.
\newblock \emph{Cognitive Neuroscience}, \emph{6}(4), 187--214.

\bibitem[{Tschantz et~al.(2020)Tschantz, Baltieri, Seth, \& Buckley}]{tschantz2020scaling}
Tschantz, A., Baltieri, M., Seth, A.~K., \& Buckley, C.~L. (2020).
\newblock Scaling Active Inference.
\newblock In \emph{Proceedings of the 2020 International Joint Conference on Neural Networks (IJCNN)} (pp. 1--8). IEEE.


\bibitem[{Millidge(2020)}]{millidge2020deep}
Millidge, B. (2020).
\newblock Deep active inference as variational policy gradients.
\newblock \emph{Journal of Mathematical Psychology}, \emph{96}, 102348.

\bibitem[{Friston et~al.(2009)}]{friston2009reinforcement}
Friston, K.~J., Daunizeau, J., \& Kiebel, S. (2009).
\newblock Reinforcement Learning or Active Inference?
\newblock \emph{PLoS ONE}, \emph{4}(7), e6421.

\bibitem[{Houlsby et~al.(2011)}]{houlsby2011bayesian}
Houlsby, N., Husz{\'a}r, F., Ghahramani, Z., \& Lengyel, M. (2011).
\newblock Bayesian Active Learning for Classification and Preference Learning.
\newblock \emph{arXiv preprint arXiv:1112.5745}.

\bibitem[{Gal et~al.(2017)}]{gal2017deep}
Gal, Y., Islam, R., \& Ghahramani, Z. (2017).
\newblock Deep Bayesian Active Learning with Image Data.
\newblock \emph{Proceedings of the 34th International Conference on Machine Learning}, \emph{1183--1192}.

\bibitem[{Kozachenko and Leonenko(1987)}]{kozachenko1987sample}
Kozachenko, L.~F., \& Leonenko, N.~N. (1987).
\newblock Sample Estimate of the Entropy of a Random Vector.
\newblock \emph{Problems of Information Transmission}, \emph{23}(2), 95--101.

\bibitem[{Millidge(2019)}]{millidge2019deep}
Millidge, B. (2019).
\newblock Deep active inference as variational policy gradients.
\newblock \emph{Journal of Mathematical Psychology}, \emph{96}, 102348.






\end{thebibliography}

\end{document}